\documentclass{article}
\usepackage{spconf,amsmath,graphicx}
\usepackage{amsmath}
\usepackage{amssymb}
\usepackage{enumitem}
\usepackage{booktabs}
\setlist{nosep, leftmargin=14pt}
\usepackage{float}
\usepackage{multirow}
\usepackage{array}
\usepackage{mwe}
\usepackage{placeins}
\newcommand{\best}[1]{\textbf{#1}}
\usepackage{hyperref}
\usepackage{makecell}
\usepackage[percent]{overpic}
\title{FocusSDF: Boundary-Aware Learning for Medical Image Segmentation via Signed Distance Supervision}

\name{Muzammal Shafique$^{1}$, Nasir Rahim$^{1}$, Jamil Ahmad$^{2}$, Mohammad Siadat$^{3}$, Khalid Malik$^{1}$, Ghaus Malik$^{4}$}
\address{
$^{1}$College of Innovation and Technology (CIT), University of Michigan-Flint, Flint, MI, USA \\
$^{2}$College of Information Technology (CIT), United Arab Emirates University, Al Ain, Abu Bhabi, UAE \\
$^{3}$Department of Computer Science and Engineering, Oakland University, Rochester Hills, MI, USA \\
$^{4}$Executive Vice-Chair at Department of Neurosurgery, Henry Ford Health System, Detroit, MI, USA
}
\begin{document}
\maketitle
\begin{abstract}
Segmentation of medical images constitutes an essential component of medical image analysis, providing the foundation for precise diagnosis and efficient therapeutic interventions in clinical practices. Despite substantial progress, most segmentation models do not explicitly encode boundary information; as a result, making boundary preservation a persistent challenge in medical image segmentation. To address this challenge, we introduce FocusSDF, a novel loss function based on the signed distance functions (SDFs), which redirects the network to concentrate on boundary regions by adaptively assigning higher weights to pixels closer to the lesion or organ boundary, effectively making it boundary aware. To rigorously validate FocusSDF, we perform extensive evaluations against five state-of-the-art medical image segmentation models, including the foundation model MedSAM, using four distance-based loss functions across diverse datasets covering cerebral aneurysm, stroke, liver, and breast tumor segmentation tasks spanning multiple imaging modalities. The experimental results consistently demonstrate the superior performance of FocusSDF over existing distance transform based loss functions. Code is available at 
\href{https://github.com/muzammalshafique/FocusSDF}{\text{GitHub}}.
\end{abstract}\vspace{-2mm}
\vspace{-4mm}
\section{Introduction}
\begin{figure}[t]
  \centering
  \includegraphics[width=\columnwidth]{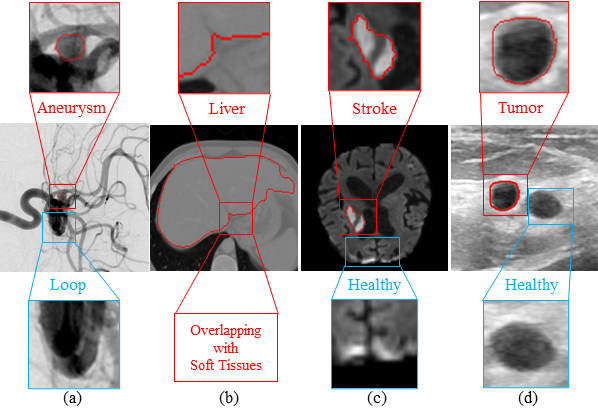}
  \vspace{-10mm}
  \caption{\small{Challenges in medical image segmentation across different modalities: (a) DSA: a partially hidden aneurysm behind the cluttered arteries and loops having similar dye contrasts, (b) CTA: an overlapping soft tissues with liver due to low tissue contrast, (c) MRI: an irregular boundaries and identical contrast of stroke and healthy tissue, and (d) Ultrasound: similar tumor and healthy tissue structure in breast.}}
  \label{fig:abstract_image}\vspace{-2mm}
\end{figure}

\vspace{-4mm}
\label{sec:intro}
Accurate delineation of regions of interest (ROI) from surrounding vasculature or tissues is fundamental for developing reliable clinical management strategies\cite{8272030}. However, constructing a unified medical image segmentation framework that can robustly generalize across both vascular and non-vascular pathological and non-pathological structures, while maintaining consistent performance across diverse imaging modalities, remains a major challenge. Even within the cardio and cerebrovascular domains, segmentation tasks exhibit domain-specific and modality-dependent complexities. For instance, cerebral aneurysm segmentation in Digital Subtraction Angiography (DSA) must contend with extremely small, irregular lesions embedded within dense, overlapping arterial networks, where aneurysmal sacs often share identical dye intensities with surrounding vessels. Conversely, ischemic stroke segmentation in MRI is hindered by diffuse and temporally evolving lesion boundaries. Extending segmentation to non-vascular diseases/organs or alternative imaging modalities further amplifies these challenges. For example, Computed Tomography Angiography (CTA) based liver segmentation suffers from heterogeneous contrast phases and adjacent organ interference, while breast tumor segmentation in ultrasound is degraded by speckle noise and operator-dependent variability. Fig.~\ref{fig:abstract_image} emphasises both disease as well as modality-specific challenges in medical image segmentation.  When optimized solely using region-centric objectives, even strong backbones such as UNet~\cite{ronneberger2015u}, UNet++~\cite{10.1007/978-3-030-00889-5_1}, SwinUNetR~\cite{hatamizadeh2021swin}, and TransUNet~\cite{CHEN2024103280} often produce discontinuous contours, particularly in small or low-contrast structures.
Moreover, foundation models such as MedSAM~\cite{MedSAM}, despite extensive pretraining, exhibit limited generalization in segmentation even when fine-tuned on complex DSA aneurysm data, where densely overlapping vasculature and low-contrast boundaries make precise localization particularly challenging. The choice of loss function is pivotal in advancing medical image segmentation performance, as task-specific formulations such as Dice, focal, and boundary-aware losses effectively mitigate class imbalance and enhance structural precision. Motivated by this, we aim to develop a unified loss function capable of generalizing across diverse diseases, anatomical structures, and imaging modalities.

In medical image segmentation, region-based losses such as Dice and Tversky \cite{10.1007/978-3-319-67389-9_44} promote region-level overlap and alleviate class imbalance, while Focal loss \cite{lin2017focal} down-weights easy negatives to emphasize harder examples. Despite their success, these methods operate purely on binary masks and are insensitive to fine geometric details, often producing incomplete boundaries. Several distance and boundary-aware losses have been proposed in literature. Ma et al. \cite{pmlr-v121-ma20b} summarized various methods, including the method level advancements as well as with loss functions. Karimi et al.~\cite{karimi2019reducing} minimized the Hausdorff distance to reduce large contour deviations, while Kervadec et al.~\cite{KERVADEC2021101851} introduced a boundary loss based on signed distance maps to emphasize contour localization in small structures. Xue et al.~\cite{xue2020shape} directly regressed SDFs for segmentation. More recent distance-based weighting schemes~\cite{10640515} show growing interest in geometry-aware supervision.
Conventional distance-based losses yield unstable gradients by treating all pixels equally, causing abrupt optimization updates and hence poor boundary convergence, specially in start of learning process. In contrast, FocusSDF emphasizes pixels near structural boundaries and down-weights distant ones, stabilizing gradients for smoother convergence and sharper boundary delineation. Table~\ref{tab:segmentation_comparison} summarizes comparison of FocusSDF with the proposed losses in literature, based on our experiments.

\begin{table}[!t]
  \centering
  \setlength{\tabcolsep}{4pt} 
  \caption{\small{Comparison of different distance transform and SDF based loss functions with FocusSDF. Abbreviations: H = High, M = Medium, L = Low}}
  \label{tab:segmentation_comparison}
  \small
  \begin{tabular}{lccccc}
    \toprule
    \textbf{Factors} & 
    \textbf{\cite{xue2020shape}} & 
    \textbf{\cite{karimi2019reducing}} & 
    \textbf{\cite{KERVADEC2021101851}} & 
    \textbf{\cite{10640515}} & 
    \textbf{FocusSDF} \\
    \midrule
    Boundary Awareness            & L & M & H & L & H \\
    Representation Type          & SDF & DT & SDF & DT & SDF \\
    Seg. Accuracy         & H & M & M & M & H \\
    Clinical Interpret.     & L & M & H & L & H \\
    Cross-Modality General.   & H & M & L & H & H \\
  \bottomrule
  \end{tabular}
\end{table}
   

To address these limitations, we propose \textbf{FocusSDF}, a boundary-aware loss function that leverages the SDF representation. Unlike binary mask-based losses that treat all pixels equally, FocusSDF explicitly encodes the distance of each pixel to the nearest boundary and applies an adaptive exponential weighting that emphasizes pixels close to contours. Our main contributions are as follows:
\begin{enumerate}
   

    \item We propose a novel boundary-aware loss function based on signed distance maps that adaptively weights pixel errors using exponential decay around object contours, enhancing geometric sensitivity.
    \item We propose a joint-learning strategy for implementation of FocusSDF that predicts binary masks and SDFs, enabling complementary region and boundary-level supervision.
    \item Extensive benchmarking across UNet, UNet++, SwinUNetR, TransUNet, and MedSAM on four modalities (DSA, CTA, MRI, and ultrasound), demonstrating superior boundary accuracy and generalization.
\end{enumerate}\vspace{-2mm}
\section{Methodology}\vspace{-3mm}
\label{sec:methodology}
\subsection{Signed Distance Function}
\begin{figure}[t]
  \centering
  \includegraphics[width=\columnwidth]{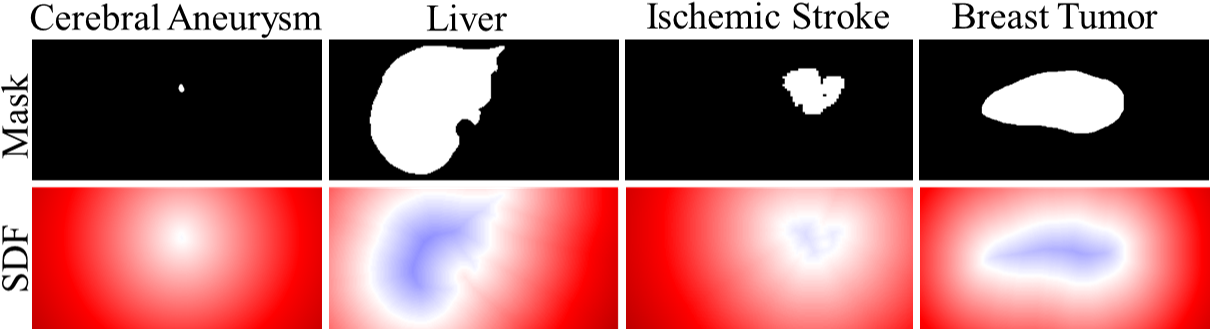}
  \vspace{-8mm}
  \caption{\small{SDF representation of Binary Masks.}}
  \label{fig:sdf_datasets}
\end{figure}

\vspace{-2mm}Let $\Omega \subset \mathbb{R}^n$ be a region with boundary $\partial \Omega$. 
For any point $\mathbf{x} \in \mathbb{R}^n$, the Euclidean distance from $\mathbf{x}$ to the boundary $\partial \Omega$ is $d(\mathbf{x}, \partial \Omega)$.
The \textit{SDF} is then defined as:\vspace{-2mm}
\begin{equation}
\phi(\mathbf{x}) =
\begin{cases}
-\,d(\mathbf{x}, \partial \Omega), & \text{if } \mathbf{x} \in \Omega, \\[2pt]
0, & \text{if } \mathbf{x} \in \partial \Omega, \\[2pt]
\;\; d(\mathbf{x}, \partial \Omega), & \text{if } \mathbf{x} \in \Omega^{c}
\end{cases}\vspace{-2mm}
\label{eq:sdf_def}
\end{equation}
Fig.~\ref{fig:sdf_datasets} shows SDF representation of binary map of different regions.
\setlength{\abovecaptionskip}{2pt}
\setlength{\belowcaptionskip}{0pt}
\vspace{-4mm}
\subsection{Proposed Loss Function Formulation}
\begin{figure}[!htbp]
  \centering
  \includegraphics[width=\columnwidth]{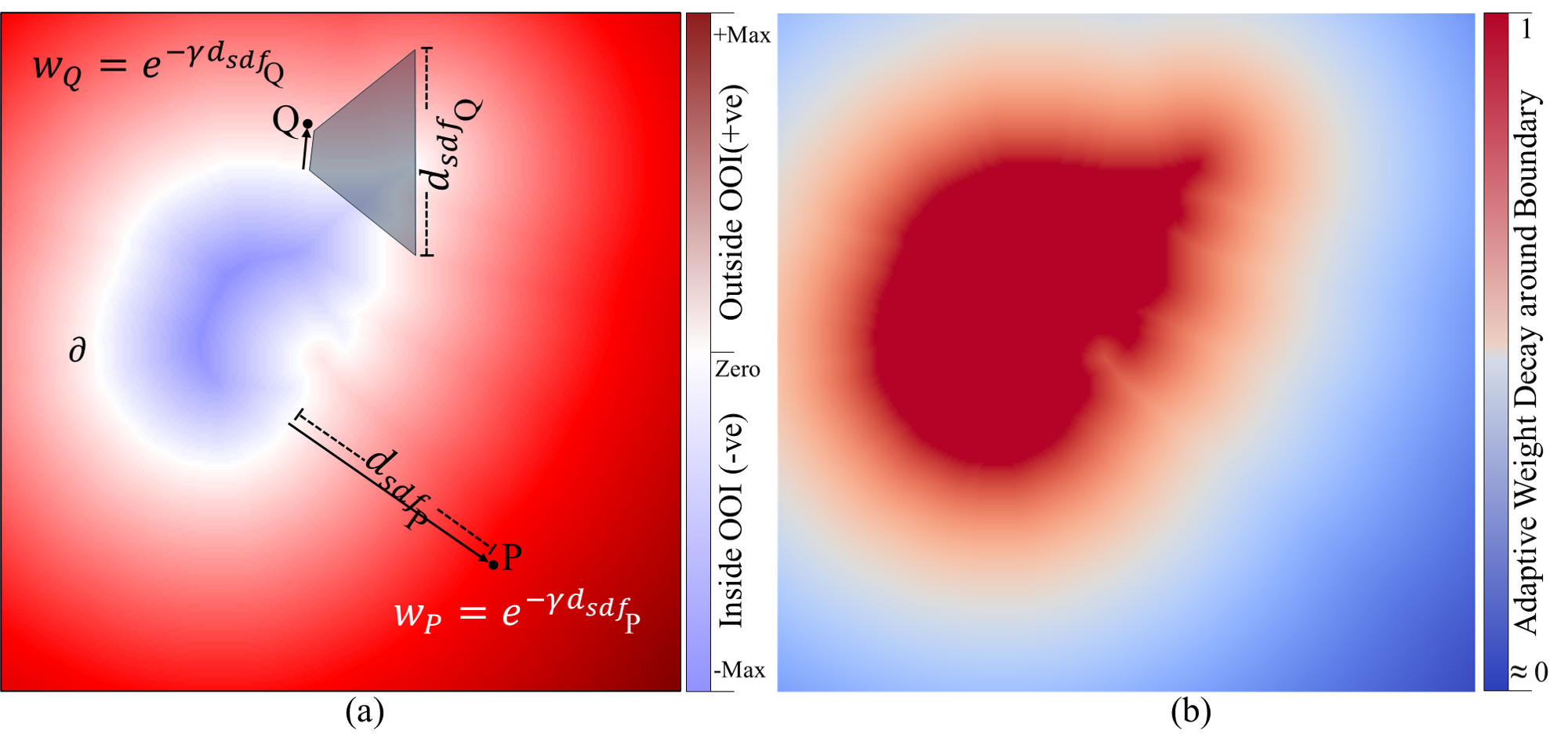}
  \vspace{-10mm}
  \caption{\small{(a) Visualization of exponential weighting that highlights boundary pixels and suppresses distant regions. (b) Spatial heatmap showing higher weights near boundaries and reduced weights in distant regions for $\gamma = 0.005$.}}
  \label{fig:loss_function_working}
\end{figure}

\vspace{-2mm}
To emphasize learning around object boundaries, we define an adaptive, distance-based weighting mechanism for SDF supervision. Unlike binary mask losses that treat all pixels equally, FocusSDF assigns higher importance to pixels near the boundary, where geometric precision is most critical.
Let $S$ and $\hat{S}$ denote the ground truth and predicted SDF maps, respectively.  
The spatial domain $\Omega$ is partitioned as:
\[
\Omega_{\text{in}} = \{\,\mathbf{i}\!:\! S_{\mathbf{i}} < 0\,\},\text{ }
\partial\Omega = \{\,\mathbf{i}\!:\! S_{\mathbf{i}} = 0\,\},\text{ }
\Omega_{\text{out}} = \{\,\mathbf{i}\!:\! S_{\mathbf{i}} > 0\,\}
\]
The proposed FocusSDF integrates the boundary-aware weighting with gradient consistency term:
\begin{equation}
\mathcal{L}_{\text{FocusSDF}} =
\frac{1}{|\Omega|}\sum_{i \in \Omega} w_i *
\big|{S}_i - \hat{S}_i\big|^p 
+ \lambda
\frac{1}{|\Omega|}\sum_{i \in \Omega}
\big|\nabla {S}_i - \nabla \hat{S}_i\big|^p,
\label{eq:focus_sdf_final}
\end{equation}\vspace{-2mm}
where,\vspace{-2mm}
\[
w_i =
\begin{cases}
1, & \text{if } \mathbf{i} \in \Omega_{\text{in}} \text{ or }\partial\Omega \\
\exp\left( -\gamma \cdot |s_{\mathbf{i}}| \right), & \text{if } \mathbf{i} \in \Omega_{\text{out}}
\end{cases}
\]
where $p \in \{1,2\}$, and $\gamma > 0$ controls the weight decay around the boundary of object and $\lambda$ balances both terms. Fig.~\ref{fig:loss_function_working} demonstrates the working of proposed weight decay phenomenon in signed distance maps.
\vspace{-2mm}
\section{Experiments}\vspace{-2mm}
In this section, we evaluate the proposed loss across diverse medical image segmentation datasets and modalities. To assess its effectiveness, the method is benchmarked against four state-of-the-art SDF-based losses using representative backbones: UNet, UNet++, SwinUNetR, TransUNet and the foundation model MedSAM, covering both convolution and transformer based paradigms.\vspace{-4mm}
\subsection{Datasets and Compared Loss Functions}\vspace{-2mm}
To evaluate the effectiveness of the proposed loss function, experiments were performed on in-house cerebral aneurysm and publicly available liver, ischemic stroke, and breast tumor segmentation datasets, corresponding to DSA, CTA, MRI, and Ultrasound modalities, respectively \cite{KAVUR2021101950, hernandez2022isles, breast_ultrasound_kaggle_aryashah2k}. For each dataset, 20\% of the patients were allocated to the test set, while 5\% of the original training split was further set aside for validation. In-house aneurysm testing dataset from Henry Ford Health System (IRB No. 11254) was made extremely complex intentionally by including most complex views and multi-aneurysm patients, in order to mimic real world clinical settings. We benchmark FocusSDF against key region and distance-based losses inlcuding Dice loss, and losses proposed in ~\cite{xue2020shape}, ~\cite{karimi2019reducing}, ~\cite{KERVADEC2021101851}, ~\cite{10640515} studies. To stabilize training process, each referenced loss was combined with the Dice loss \cite{pmlr-v121-ma20b}, and the resulting formulations are denoted as $\mathcal{L}_{1}$, $\mathcal{L}_{2}$, $\mathcal{L}_{3}$ and $\mathcal{L}_{4}$, respectively.\vspace{-4mm}
\subsection{Implementation Details and Performance Metrics}\vspace{-2mm}
All experiments were implemented in Python~3.13.1 and PyTorch~2.6.0 with CUDA~11.8 on an NVIDIA RTX~4090 Tesla GPU. Input images wrer resized to 256×256. Each model produced two outputs: binary segmentation logits and signed distance maps (SDFs), trained for 200 epochs with a batch size of 8 using AdamW (lr=$10^{-4}$). For numerical stability, ground truth SDF is standard normalized. Each loss function has been implemented to closely replicate its originally proposed formulation, to the best of our understanding. We employ three segmentation metrics of \textbf{Dice}, \textbf{IoU}, and \textbf{HD95}, to record the general segmentation performance as well as boundary awareness.\vspace{-4mm}
\subsection{Results}\vspace{-2mm}
\label{ssec:results}
\begin{table*}[t]
\centering
\caption{\small{Comparison of loss functions across datasets using UNet, UNet++, SwinUNetR, and TransUNet. HD95 is in pixels.}}
\label{tab:loss_comparison}
\small
\begin{tabular}{|l|l|ccc|ccc|ccc|ccc|}
\toprule[0.5pt]
\textbf{Dataset} & \textbf{Loss} &
\multicolumn{3}{c}{\textbf{UNet}} &
\multicolumn{3}{c}{\textbf{UNet++}} &
\multicolumn{3}{c}{\textbf{SwinUNetR}} &
\multicolumn{3}{c|}{\textbf{TransUNet}} \\
\cmidrule(lr){3-5} \cmidrule(lr){6-8} \cmidrule(lr){9-11} \cmidrule(lr){12-14}
 &  & \textbf{Dice↑} & \textbf{IoU↑} & \textbf{HD95↓} 
 & \textbf{Dice↑} & \textbf{IoU↑} & \textbf{HD95↓} 
 & \textbf{Dice↑} & \textbf{IoU↑} & \textbf{HD95↓} 
 & \textbf{Dice↑} & \textbf{IoU↑} & \textbf{HD95↓} \\
\midrule

\multirow{6}{*}{Aneurysm}
 & $\mathcal{L}_{Dice}$  & 0.535 & 0.501 & 37.0 & 0.433 & 0.343 & 39.0 & 0.554 & 0.468 & 32.2 & 0.545 & 0.467 & 30.9 \\
 & $\mathcal{L}_{1}$  & 0.535 & 0.501 & 35.0 & 0.480 & 0.447 & 37.0 & 0.554 & 0.468 & 34.8 & 0.438 & 0.407 & 36.5 \\
 & $\mathcal{L}_{2}$  & 0.553 & 0.505 & 35.5 & 0.485 & 0.455 & 34.0 & 0.535 & 0.453 & 35.2 & 0.588 & 0.477 & 34.9 \\
 & $\mathcal{L}_{3}$  & 0.496 & 0.462 & 36.0 & 0.548 & 0.487 & 34.0 & 0.422 & 0.385 & 35.2 & 0.299 & 0.297 & 38.7 \\
 & $\mathcal{L}_{4}$  & 0.570 & 0.528 & 32.0 & 0.488 & 0.418 & 31.6 & 0.442 & 0.328 & 33.2 & 0.400 & 0.383 & 36.8 \\
 & $Ours$  & \best{0.770} & \best{0.704} & \best{25.0} & \best{0.757} & \best{0.691} & \best{23.8} & \best{0.787} & \best{0.695} & \best{22.2} & \best{0.794} & \best{0.702} & \best{21.5} \\
\midrule

\multirow{6}{*}{Liver}
 & $\mathcal{L}_{Dice}$  & 0.914 & 0.872 & 21.2 & 0.925 & 0.880 & 19.0 & 0.905 & 0.874 & 20.4 & 0.922 & 0.885 & 19.1 \\
 & $\mathcal{L}_{1}$  & 0.920 & 0.876 & 21.1 & 0.914 & 0.872 & 19.9 & 0.872 & 0.818 & 22.6 & 0.907 & 0.865 & 17.2 \\
 & $\mathcal{L}_{2}$  & \best{0.951} & \best{0.925} & \best{13.8} & 0.891 & 0.833 & 27.2 & 0.862 & 0.792 & 24.1 & 0.922 & 0.885 & 19.1 \\
 & $\mathcal{L}_{3}$  & 0.317 & 0.200 & 45.4 & 0.261 & 0.155 & 42.5 & 0.307 & 0.191 & 40.3 & 0.291 & 0.181 & 38.0 \\
 & $\mathcal{L}_{4}$  & 0.898 & 0.844 & 22.8 & 0.903 & 0.857 & 20.3 & 0.902 & 0.847 & 17.6 & 0.927 & 0.892 & 14.5 \\
 & $Ours$  & 0.950 & 0.915 & 15.0 & \best{0.950} & \best{0.916} & \best{14.7} & \best{0.934} & \best{0.875} & \best{14.8} & \best{0.969} & \best{0.936} & \best{13.8} \\
\midrule

\multirow{6}{*}{Stroke}
 & $\mathcal{L}_{Dice}$  & 0.705 & 0.634 & 26.0 & 0.712 & 0.655 & 25.1 & \best{0.786} & \best{0.700} & \best{14.2} & 0.755 & 0.684 & 24.3 \\
 & $\mathcal{L}_{1}$  & 0.698 & 0.611 & 26.5 & 0.698 & 0.567 & 25.9 & 0.702 & 0.597 & 17.7 & 0.708 & 0.599 & 15.4 \\
 & $\mathcal{L}_{2}$  & 0.701 & 0.620 & 26.2 & 0.702 & 0.617 & 25.7 & 0.655 & 0.546 & 20.3 & 0.689 & 0.576 & 18.4 \\
 & $\mathcal{L}_{3}$  & 0.346 & 0.246 & 48.2 & 0.365 & 0.272 & 46.4 & 0.325 & 0.258 & 41.3 & 0.206 & 0.118 & 38.3 \\
 & $\mathcal{L}_{4}$  & 0.683 & 0.602 & 25.4 & 0.695 & 0.606 & 24.8 & 0.684 & 0.575 & 20.2 & 0.706 & 0.597 & 17.7 \\
 & $Ours$  & \best{0.758} & \best{0.675} & \best{20.8} & \best{0.756} & \best{0.667} & \best{18.9} & 0.775 & 0.698 & 15.5 & \best{0.754} & \best{0.666} & \best{15.3} \\
\midrule

\multirow{6}{*}{\makecell{Breast\\Tumor}}
 & $\mathcal{L}_{Dice}$  & 0.684 & 0.635 & 27.0 & 0.715 & 0.642 & 25.8 & 0.651 & 0.584 & 24.9 & 0.696 & 0.629 & 22.9 \\
 & $\mathcal{L}_{1}$  & 0.756 & 0.688 & 24.6 & 0.747 & 0.647 & 25.1 & 0.612 & 0.533 & 29.6 & 0.626 & 0.545 & 29.7 \\
 & $\mathcal{L}_{2}$  & 0.747 & 0.664 & 22.4 & 0.758 & 0.665 & 24.1 & 0.606 & 0.571 & 28.1 & 0.499 & 0.399 & 30.3 \\
 & $\mathcal{L}_{3}$  & 0.384 & 0.310 & 52.2 & 0.299 & 0.226 & 50.5 & 0.611 & 0.181 & 45.0 & 0.388 & 0.329 & 42.3 \\
 & $\mathcal{L}_{4}$  & 0.656 & 0.584 & 29.7 & \best{0.817} & \best{0.721} & \best{18.7} & 0.659 & 0.581 & 25.4 & 0.679 & 0.599 & 20.6 \\
 & $Ours$  & \best{0.791} & \best{0.704} & \best{20.2} & 0.802 & 0.732 & 19.9 & \best{0.725} & \best{0.647} & \best{22.3} & \best{0.765} & \best{0.682} & \best{21.8} \\
\bottomrule
\end{tabular}
\end{table*}

\begin{figure}[t]
      \centering
      \includegraphics[width=\columnwidth]{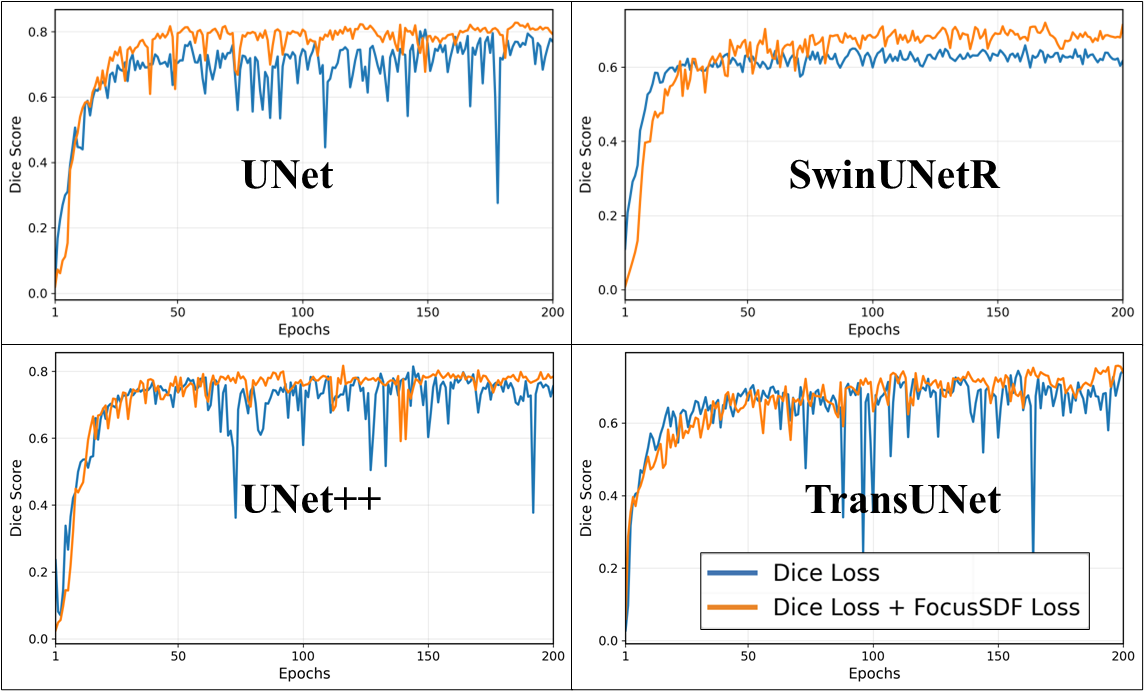}
      \vspace{-8mm}
      \caption{\small{Validation performance of models on aneurysm segmentation dataset.}}
      \label{fig:per_loss_curves}
    \end{figure}

\begin{figure*}[!htbp]
  \centering
  \includegraphics[width=1\textwidth]{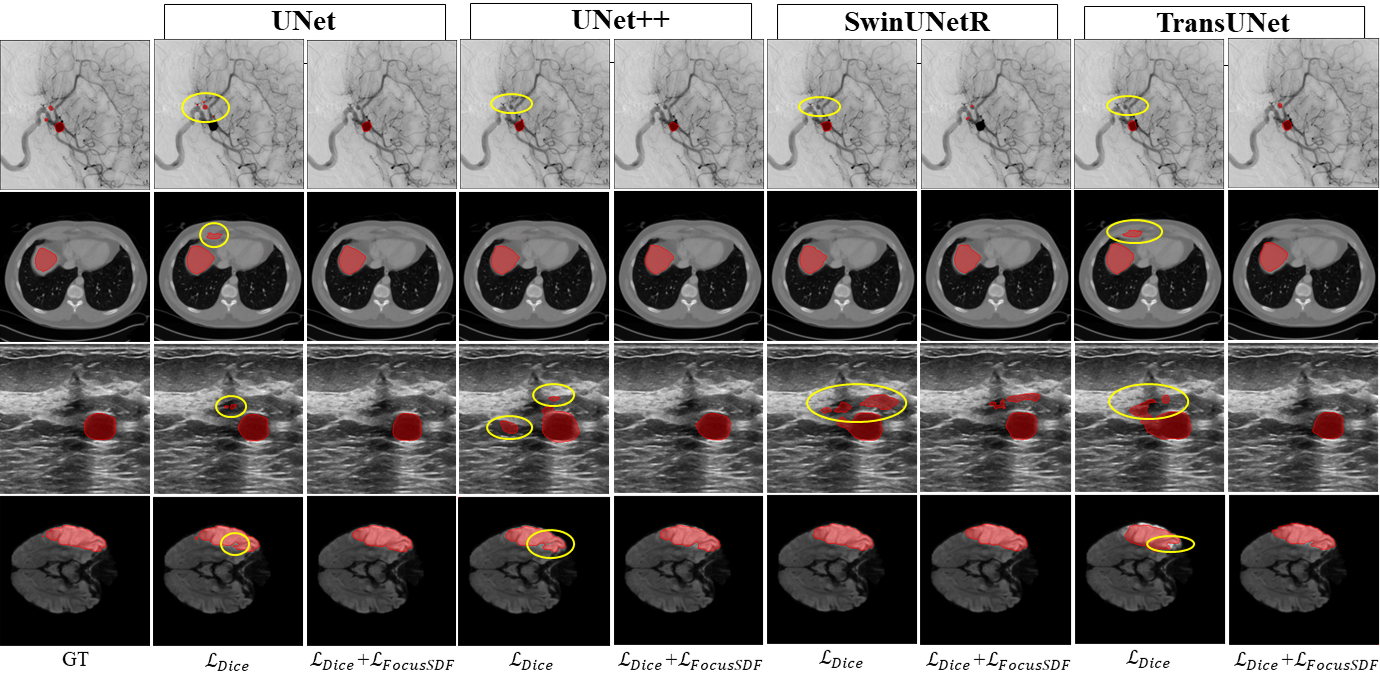}\vspace{-2mm}
  \caption{Qualitative comparison of segmentation outputs across datasets. Incorporating the proposed FocusSDF loss alongside the Dice loss yields superior boundary precision and structural continuity preservation compared to using the Dice loss alone. Mispredictions with dice loss are encircled as shown.}
  \label{fig:big_hh_o}
\end{figure*}

Table~\ref{tab:loss_comparison} summarizes the quantitative as well as Fig.~\ref{fig:big_hh_o}presents the qualitative results of the proposed FocusSDF loss. FocusSDF achieves relatively superior overlap and boundary accuracy, especially in challenging case such as cerebral aneurysm. Its consistent performance across both CNN and transformer-based models highlights its generality and robustness in handling diverse anatomical structures and imaging modalities. Fig.~\ref{fig:per_loss_curves} highlights the convergence and stability behavior of the FocusSDF on aneurysm dataset across models.

To complement the results in Table~\ref{tab:loss_comparison}, we fine-tuned the foundation model \textbf{MedSAM} on the aneurysm dataset using dice loss. Despite fine-tuning, it achieved a modest dice score of 0.3340 on DSA images. To our best knowledge, possible reasons of this performance degade is due to highly intricate vascular structure and overlapping dye intensities of aneurysms, loops and arterial network, coupled with the absence of DSA modality in its pretraining and fine-tuning datasets. Nevertheless, its one-shot generalization on CT liver, breast tumor ultrasound, and ischemic stroke MRI datasets yielded moderate dice scores of 0.8941, 0.7538, and 0.4322, respectively, reflecting better transferability from modalities that were well-represented during it pretraining.

\vspace{-4mm}
\section{Conclusion and Future Work}\vspace{-2mm}
In this work, we introduced FocusSDF, a boundary-aware loss function that adaptively prioritizes pixels based on their proximity to the object boundary. By combining adaptive boundary weighting with a gradient-consistency term, our formulation encourages precise boundary delineation while maintaining smooth spatial transitions across the segmentation map. Experimental results across multiple medical datasets confirm that FocusSDF improves both overlap metrics and boundary accuracy, particularly for anatomically complex regions such as aneurysm segmentation in DSA, and enhancing the overall segmentation performance on other modalities too.

Future work will extend this framework to 3D volumetric and multi-class segmentation for improved geometric consistency.

{
\bibliographystyle{IEEEbib}
\bibliography{strings,refs}
}

\end{document}